\documentclass{article}
\usepackage{spconf,amsmath,graphicx,hyperref, amsfonts}
\usepackage{amssymb}
\usepackage{algorithm,algpseudocode}
\usepackage{booktabs} 
\usepackage{array} 
\usepackage{adjustbox} 
\usepackage{multirow}
\usepackage{makecell}

\title{USCTNet: A deep unfolding nuclear-norm optimization solver for physically consistent HSI reconstruction}
\name{Xiaoyang Ma, Yiyang Chai, Xinran Qu, Hong Sun\thanks{This work was supported by the National Key Research and Development Program (Grant No. 2023YFD1701001) for funding.}}
\address{Key Lab of Smart Agriculture Systems, Ministry of Education, China Agricultural University}
\begin{document}
%
\maketitle
\begin{abstract}
Reconstructing hyperspectral images (HSIs) from a single RGB image is ill-posed and can become physically inconsistent when the camera spectral sensitivity (CSS) and scene illumination are misspecified. We formulate RGB-to-HSI reconstruction as a physics-grounded inverse problem regularized by a nuclear norm in a learnable transform domain, and we explicitly estimate CSS and illumination to define the forward operator embedded in each iteration, ensuring colorimetric consistency. To avoid the cost and instability of full singular-value decompositions (SVDs) required by singular-value thresholding (SVT), we introduce a data-adaptive low-rank subspace SVT operator. Building on these components, we develop USCTNet, a deep unfolding solver tailored to HSI that couples a parameter estimation module with learnable proximal updates. Extensive experiments on standard benchmarks show consistent improvements over state-of-the-art RGB-based methods in reconstruction accuracy. Code: \url{https://github.com/psykheXX/USCTNet-Code-Implementation.git}
\end{abstract}


\section{Introduction}
Recovering a hyperspectral image (HSI) $Y\!\in\!\mathbb{R}^{B\times N}$ from a single RGB $X\!\in\!\mathbb{R}^{3\times N}$ is highly ill-posed, and colorimetric consistency often breaks under unseen cameras or illuminations. HSIs exhibit prominent low-rank structure across bands~\cite{wang2017hyperspectral}, motivating nuclear-norm regularization as a principled, physics-compatible prior~\cite{zhang2024t2lr}. However, its proximal operator, SVT, requires repeated SVDs, leading to heavy computation and unstable gradients during deep training~\cite{zhang2024differentiable}.

\begin{figure}[t]
    \centering
    \includegraphics[width=\columnwidth]{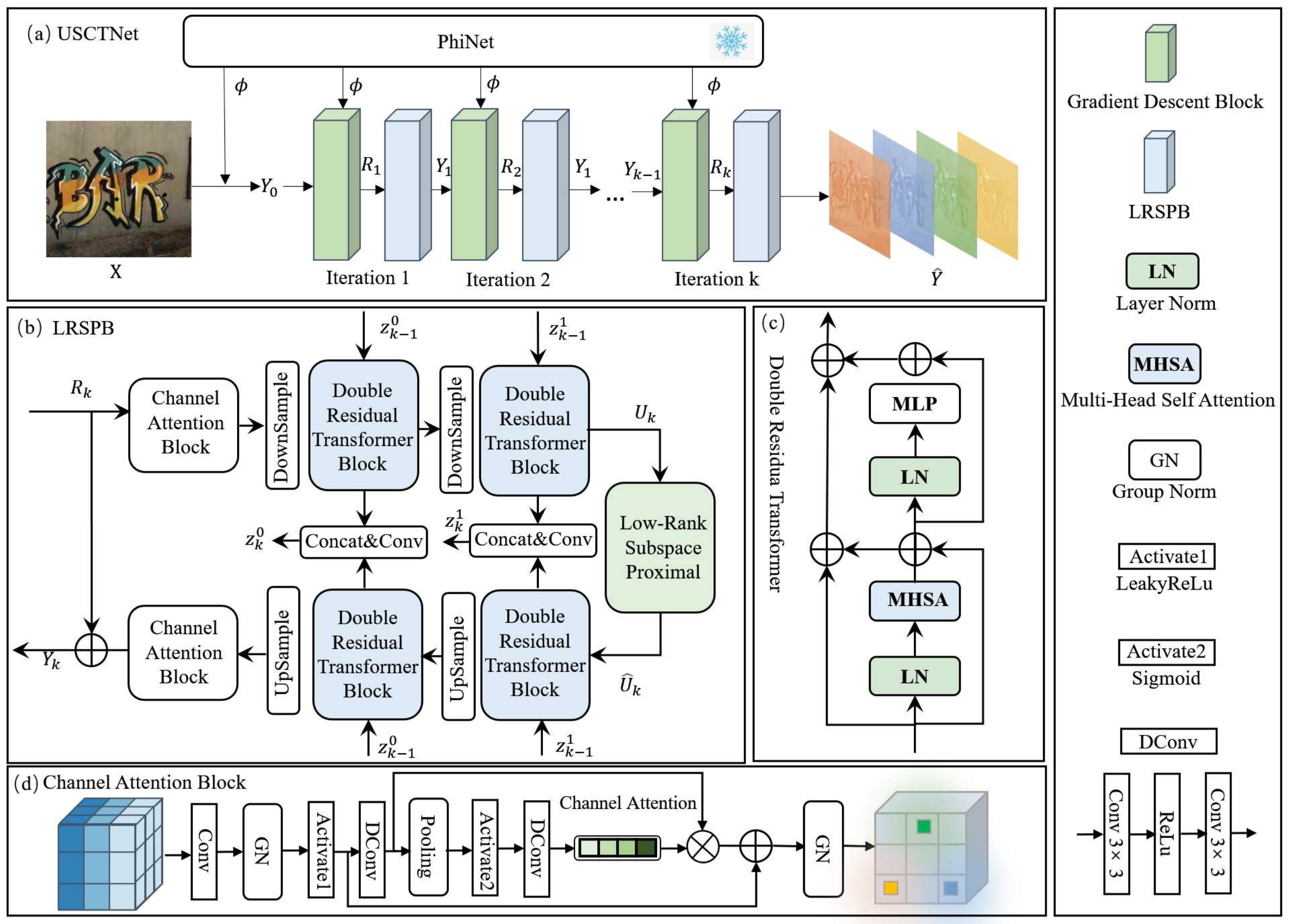}
    \vspace{-2mm}
    \caption{Illustration of the proposed USCTNet framework.}
    \label{fig.1}
\end{figure}

We address these challenges by formulating RGB-to-HSI reconstruction as a physics-grounded inverse problem and by developing \textbf{USCTNet}, an unfolding solver that embeds imaging physics and a low-rank prior. Let $S\!\in\!\mathbb{R}^{3\times B}$ denote the camera spectral sensitivity matrix and $\ell\!\in\!\mathbb{R}^{B}$ the scene-illumination spectrum; then
\begin{equation}
\label{eq:forward}
X = \Phi Y,\qquad \Phi = S\,\operatorname{Diag}(\ell).
\end{equation}
USCTNet explicitly estimates $\Phi$ to drive a physics-guided gradient step. In a learnable transform domain, a \textbf{Low-Rank Subspace Proximal (LRSP)} enforces the nuclear-norm prior by replacing full-SVD SVT with learned subspace soft-thresholding, improving stability and efficiency.

\textbf{Contributions.}\quad
(1) We propose USCTNet, a physics-grounded unfolding solver that casts RGB-to-HSI as a nuclear-norm-regularized inverse problem in a learnable transform domain. 
(2) We explicitly estimate the forward operator $\Phi$ and use it to drive data-fidelity updates, ensuring colorimetric consistency and improving generalization across cameras and illuminants. 
(3) We introduce LRSP, a learnable soft-threshold proximal applied to a data-driven subspace, avoiding full SVDs and stabilizing training. 
(4) USCTNet achieves SOTA results on standard benchmarks across reconstruction accuracy, perceptual quality and color consistency.

\section{Related Works}
\noindent\textbf{1) Spectral Reconstruction from RGB.}
Early CNN or Transformer-based methods learn RGB-to-HSI mappings by exploiting spatial–spectral correlations and achieve strong accuracy~\cite{cai2022mst++,li2020adaptive}. However, the absence of explicit physics often leads to color artifacts and inconsistent RGB reproduction under domain shift~\cite{huo2024learning}. Physics-consistent formulations that incorporate CSS constraints improve color fidelity~\cite{lin2020physically,huo2024learning}, yet they depend on network-predicted null-space components and can amplify noise through pseudoinverse projection. Moreover, few methods leverage structural priors. Given the well-documented low-rank structure of HSIs, nuclear-norm regularization is a natural prior.

\noindent\textbf{2) Deep unfolding Networks with Low-Rank Priors.}
Deep unfolding networks (DUNs) recast iterative optimization as learnable modules that leverage efficient proximal operators~\cite{wang2024ufc,zhang2018ista}. Recent efforts unfold nuclear-norm optimization via transformed or expanded representations~\cite{zhang2024t2lr,zhang2025jotlasnet}, but most still rely on SVDs, which are impractical at HSI scale and can destabilize training—motivating subspace-aware, full-SVD-free surrogates.

\noindent\textbf{3) Full-SVD-Free Proximal Mappings.}
SVT entails repeated full SVDs and becomes prohibitive for high-dimensional HSIs. Factorization methods approximate the nuclear norm with low-rank factors~\cite{xu2017low}; randomized subspace projections reduce complexity by operating in compact bases~\cite{halko2011RandQB}; and convex–concave SVD-free iterations require only leading components~\cite{xiao2017svd}. These insights motivate differentiable, learnable subspace proximal mappings suited for end-to-end training.

\section{Method}

\subsection{Notation and Problem Setup}
Let $X\!\in\!\mathbb{R}^{3\times N}$ denote the RGB observation and $Y\!\in\!\mathbb{R}^{B\times N}$ the target HSI ($N$ pixels, $B$ bands). The forward operator $\Phi$ follows Eq.\eqref{eq:forward}. The analysis–synthesis transforms are $\mathcal{T}_{\theta}:\mathbb{R}^{B\times N}\!\to\!\mathbb{R}^{d\times n}$ and $\mathcal{T}^\dagger_{\theta}:\mathbb{R}^{d\times n}\!\to\!\mathbb{R}^{B\times N}$, acting primarily along the spectral dimension. We denote by $d$ the transform feature size, by $n$ the vectorized spatial size, and by $r\!\ll\!\min\{d,n\}$ the target subspace rank.

We recover $Y$ via the regularized inverse problem
\begin{equation}
\hat{Y}
\;=\;
\arg\min_{Y\in\mathbb{R}^{B\times N}}
\frac{1}{2}\,\|\Phi\,Y-X\|_F^2
\;+\;\lambda\,\big\|\mathcal{T}_{\theta}(Y)\big\|_* .
\label{eqn.2}
\end{equation}

\begin{table*}[t]
\centering
\begingroup
\footnotesize
\setlength{\tabcolsep}{6pt}
\renewcommand{\arraystretch}{1}
\begin{tabular}{l|c|c|ccc|ccc}
\hline
Method & Params (M) & FLOPs (G)
& \multicolumn{3}{c|}{\makecell{ARAD-1K Real\\ \footnotesize PSNR$\uparrow$ / SSIM$\uparrow$ / SAM$\downarrow$}}
& \multicolumn{3}{c}{\makecell{ICVL \\ \footnotesize PSNR$\uparrow$ / SSIM$\uparrow$ / SAM$\downarrow$}} \\
\hline
AWAN (CVPRW’20)~\cite{li2020adaptive}         & 4.04 & 270.61 & 31.92 & 0.901 & 6.03  & 30.52 & 0.918 & 2.92 \\
HINet (CVPR’21)~\cite{chen2021hinet}          & 5.21 & 31.04  & 32.71 & 0.916 & 5.10  & 30.65 & 0.967 & 2.89 \\
MPRNet (CVPR’21)~\cite{zamir2021multi}        & 3.62 & 101.59 & 33.83 & 0.930 & 5.46  & 29.97 & 0.958 & 3.22 \\
Restormer (CVPR’22)~\cite{zamir2022restormer} & 15.11 & 93.77 & 33.55 & 0.927 & 5.35  & 30.74 & 0.964 & 2.66 \\
MST-L (CVPR’22)~\cite{cai2022mask}           & 2.45 & 32.07 & 34.17 & 0.933 & 5.30  & 30.85 & 0.969 & 3.13 \\
MST++ (CVPRW’22)~\cite{cai2022mst++}          & 1.62 & \textbf{23.05} & 34.36 & 0.927 & 5.10  & 31.24 & 0.968 & 3.05 \\
CESST (AAAI’24)~\cite{yang2024hyperspectral}  & \textbf{1.54} & 90.18 & 35.19 & 0.932 & 5.73  & 33.16 & 0.978 & \textbf{2.64} \\
\hline
\textbf{USCTNet (ours)} & 15.10 & 331.72 
& \textbf{36.56} & \textbf{0.941} & \textbf{4.69} 
& \textbf{35.13} & \textbf{0.979} & 2.79 \\
\hline
\end{tabular}
\caption{Quantitative comparison with state-of-the-art methods on ARAD-1K Real and ICVL. The best results are in bold.}
\label{tab:sota_comparison}
\endgroup
\end{table*}

\subsection{Unfolding Architecture}
$\Phi$ is estimated by a network and shared across stages. The step size $\eta_k\!>\!0$ stabilizes gradient updates. The analysis operator $\mathcal{T}_{\theta}$ converts the domain of HSIs. $\mathrm{LRSP}(\cdot;\tau_k)$ denotes the learnable low-rank subspace proximal with temperature schedule $\tau_k$. Finally, the synthesis operator $\mathcal{T}^\dagger_{\theta}$ converts it back to the original domain. The module $\mathcal{M}$ maintains a memory state $Z_k$ to provide coarse spectral context. The specific algorithm flow of USCTNet is shown below:
\begin{algorithm}
\caption{USCTNet: one forward pass with $K$ unfolding stages}
\centering
\resizebox{0.9\linewidth}{!}{%
\begin{minipage}{\linewidth}
\begin{algorithmic}[1]
 \Require RGB $X$, stages $K$, step sizes $\{\eta_k>0\}$, memory $Z_0$, init $Y_0$
\State $\Phi \leftarrow \texttt{PhiNet}(X)$ \Comment{stage-shared forward operator}
\For{$k=1,\dots,K$}
    \State $R_k \leftarrow Y_{k-1}-\eta_k\,\Phi^{\top}(\Phi Y_{k-1}-X)$ \Comment{physics-guided GD step}
    \State $U_k \leftarrow \mathcal{T}_{\theta}(R_k; Z_{k-1})$
    \State $\widehat U_k \leftarrow \mathrm{LRSP}(U_k; \tau_k)$ \Comment{low-rank subspace proximal}
    \State $Y_k \leftarrow \mathcal{T}^\dagger_{\theta}(\widehat U_k)$
    \State $Z_k \leftarrow \mathcal{M}(Z_{k-1}, U_k)$ \Comment{cross-stage memory update}
\EndFor
\State \Return $Y_K$
\end{algorithmic}
\end{minipage}}
\label{alg:1}
\end{algorithm}

\subsection{Low-Rank Subspace Proximal Mapping}
\label{sec.LRSP}
We design a learnable proximal operator that performs subspace SVT to avoid full-size SVD. At stage $k$, the input is $U_k\!\in\!\mathbb{R}^{d\times n}$. $r\!\ll\!\min\{d,n\}$ is the target rank, $T$ the number of inner steps, $\kappa$ the column budget, and $s$ the number of Gaussian probes. We describe one inner iteration $t$ within stage $k$.

\textbf{(i) Differentiable selection \& sketch.}
Compute nonnegative importances $g\!\in\!(0,1)^n$ via a gated convolution and column pooling. Embed each column by $f_{\mathrm{emb}}:\mathbb{R}^{d}\!\to\!\mathbb{R}^{m}$ and score with a query $\mathbf{q}\!\in\!\mathbb{R}^{m}$:
\begin{equation}
s_i=\langle f_{\mathrm{emb}}(U_k[:,i]),\mathbf{q}\rangle,\quad 
\tilde{s}_i=\frac{s_i-s_{(\kappa+1)}}{\tau_t},
\end{equation}
where $s_{(\kappa+1)}$ is stop-gradient and $\tau_t\!>\!0$ is a temperature. The soft top-$\kappa$ weights are
\begin{equation}
\big[\mathrm{SoftTop}_{\kappa}(s;\tau_t)\big]_i=\frac{\mathrm{softplus}(\tilde{s}_i)}{\sum_j \mathrm{softplus}(\tilde{s}_j)}.
\end{equation}
Define the selector $\Omega_t=\operatorname{Diag}(g)\,\mathrm{SoftTop}_{\kappa}(s;\tau_t)\in\mathbb{R}^{n\times \kappa}$ and the orthonormal subspace
\begin{equation}
Q_t \;=\; \operatorname{orth}(U_k\,\Omega_t)\in\mathbb{R}^{d\times r}.
\label{eqn.5}
\end{equation}

\textbf{(ii) Adaptive probing \& threshold schedule.}
With $\Xi\!\sim\!\mathcal{N}(0,1)^{n\times s}$ and $G=\operatorname{Diag}(g)\,\Xi$, estimate the residual ratio, which is used as fusion weight in Eqn.\ref{eqn.11}:
\begin{equation}
\tilde{\rho}_t=\frac{\|(I-Q_tQ_t^\top)U_k G\|_F}{\|U_k G\|_F+\varepsilon}\in(0,1).
\end{equation}
Refine and predict the threshold increment using a lightweight MLP:
\begin{equation}
(\hat{\rho}_t,\Delta\beta_t)=\mathrm{MLP}_\eta\!\left(\left[\mathrm{SparsePool}(U_k;\Omega_t),\tilde{\rho}_t\right]\right),
\end{equation}
where $\mathrm{SparsePool}(U_k;\Omega_t)=\sum_{i=1}^n\Omega_t(i)\,U_k[:,i]\in\mathbb{R}^d$. Use an exponential schedule with cumulative soft threshold
\begin{equation}
\tau_t=\max(\tau_{\min},\,\tau_0\gamma^{t-1}),\qquad
\beta_{t+1}=\beta_t+\Delta\beta_t,\;\;\beta_1>0.
\label{eqn.8}
\end{equation}

\begin{figure*}[!t]
    \centering
    \includegraphics[width=\linewidth]{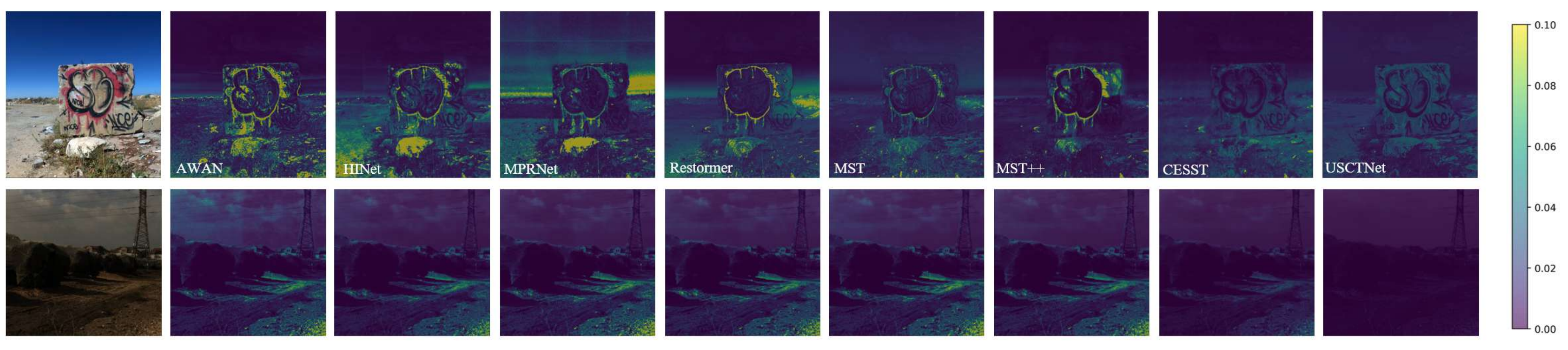}
    \caption{Mean-squared-error (MSE) maps on validation subsets of two datasets, computed along the spectral dimension.}
    \label{fig.2}
\end{figure*}

\textbf{(iii) Subspace proximal \& fusion.}
At iteration $t$, $U_k$ is projected onto the subspace spanned by $Q_t$ (Eqn.~\ref{eqn.5}) to form the compact matrix $B_t$. A depthwise-separable module, $\operatorname{ConvSVT}_{\psi}(\cdot)$, is then applied as a data-adaptive surrogate for singular-value shrinkage. Its output is combined with the identity via the gate $\alpha_t=\sigma(\beta_t)\in(0,1)$, where $\beta_t$ denotes the threshold increment (Eqn.~\ref{eqn.8}). The result is finally mapped back to the ambient space.
\begin{equation}
\widetilde{B}_{t}=(1-\alpha)B_t+\alpha(\operatorname{ConvSVT}_{\psi}\!\big(B_t)), \widehat{U}_{k,t} = Q_t\,{B}_t.
\end{equation}
Fuse proposals with residual-aware soft weights:
\begin{equation}
w_t=\frac{\exp(-\nu\hat{\rho}_t)}{\sum_{s=1}^{T}\exp(-\nu\hat{\rho}_s)},\ \nu>0,\qquad
\widehat{U}_k=\sum_{t=1}^{T} w_t\,\widehat{U}_{k,t}.
\label{eqn.11}
\end{equation}

\textbf{Complexity \& differentiability.}
Per inner step: subspace sketch $\mathcal{O}(dn)$; $\operatorname{orth}(\cdot)$ to rank $r$: $\mathcal{O}(dr^2)$; projection $\mathcal{O}(dnr)$; thin SVD on $r\times n$: $\mathcal{O}(r^2n)$. Overall $\mathcal{O}(dnr+r^2n)$ with $r\!\ll\!n$, avoiding full SVD $\mathcal{O}(dn\min\{d,n\})$. All components are differentiable: softplus with $\tau_t$ relaxes top-$\kappa$, and SVD backprop follows standard autodiff with soft shrinkage.

\subsection{Algorithm Network Implementation}
We adopt the architecture of \cite{yang2024hyperspectral} to predict the forward operator $\Phi$, referred to as \textbf{PhiNet}. PhiNet is pre-trained and kept frozen during backbone training to avoid instability from joint optimization. To prevent data leakage in our experiments, we train PhiNet using the same dataset train split as in Sec.~\ref{experiments}. The outer branch replaces the traditional analysis–synthesis transform ($\mathcal{T}_{\theta}$–$\mathcal{T}^\dagger_{\theta}$) with the encoder–decoder of ~\cite{cai2022mst++}, customized for HSI data to strengthen cross-band representation. To further enhance stability, convolution-guided channel attention (\textbf{CAB}; see Fig.~\ref{fig.1}D) is inserted at both the input and output, and Transformer layers are equipped with dual residual transformer blocks (\textbf{DRTBs}; see Fig.~\ref{fig.1}C) to suppress gradient oscillations in high-dimensional features. These modules, together with LRSP, constitute the \textbf{LRSPB} (see Fig.~\ref{fig.1}B). Finally, a concatenation followed by a convolutional layer learns the auxiliary iterative update term $\mathcal{M}$. The backbone is trained end-to-end with an $\ell_1$ reconstruction loss.

\section{Experiments and Results}
\label{experiments}

\textbf{Implementation Details.} We implement the model in PyTorch. Inputs are normalized to $[0,1]$ and cropped into $128\times128$ patches with a stride of $8$. Data augmentation includes random flips and rotations. We train with Adam (initial learning rate $10^{-5}$, $\beta_1\!=\!0.9$, $\beta_2\!=\!0.999$) using a cosine annealing schedule for 300 epochs and a batch size of 8. All experiments are conducted on an NVIDIA A100 (40\,GB).

\textbf{Datasets.} We evaluate USCTNet on ARAD-1K Real~\cite{ntiredataset} and ICVL~\cite{icvldataset}. Because ICVL lacks paired RGB images, we synthesize RGBs following \cite{hsi2rgb}. Eighteen images with inconsistent resolutions are removed, yielding 153 training and 30 test pairs.

\textbf{Quantitative Results.} We compare USCTNet with SOTA methods using PSNR, SSIM, and SAM. As shown in Table~\ref{tab:sota_comparison}, USCTNet achieves the best overall accuracy.

\begin{figure}
    \centering
    \includegraphics[width=0.9\columnwidth]{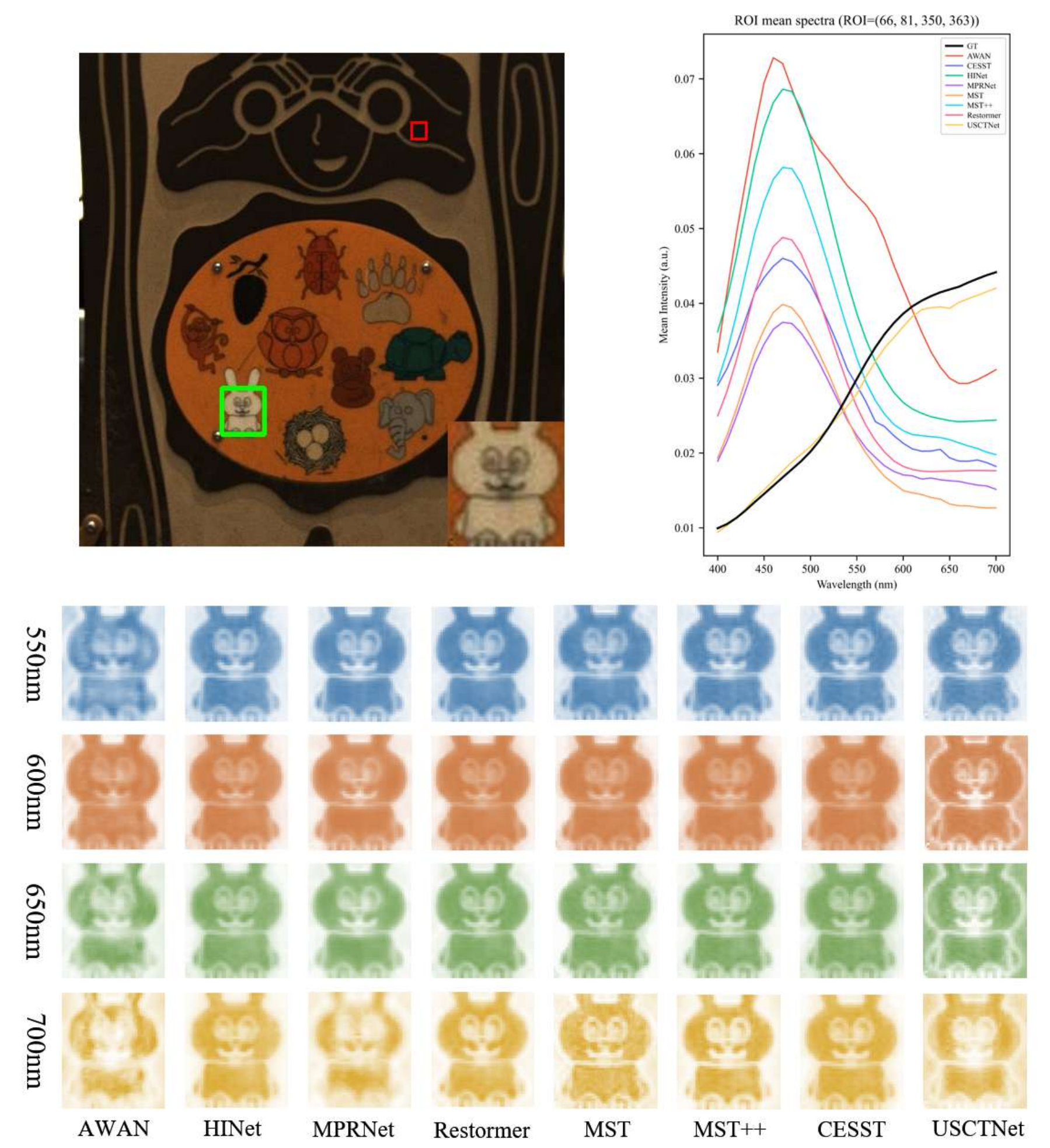}
    \caption{Visual comparison on a randomly selected ARAD-1K validation scene with four spectral channels. The spectral curves (top-right) correspond to the red-highlighted region in the RGB image.}
    \label{fig.3}
\end{figure}

\begin{figure}
    \centering
    \includegraphics[width=0.9\columnwidth]{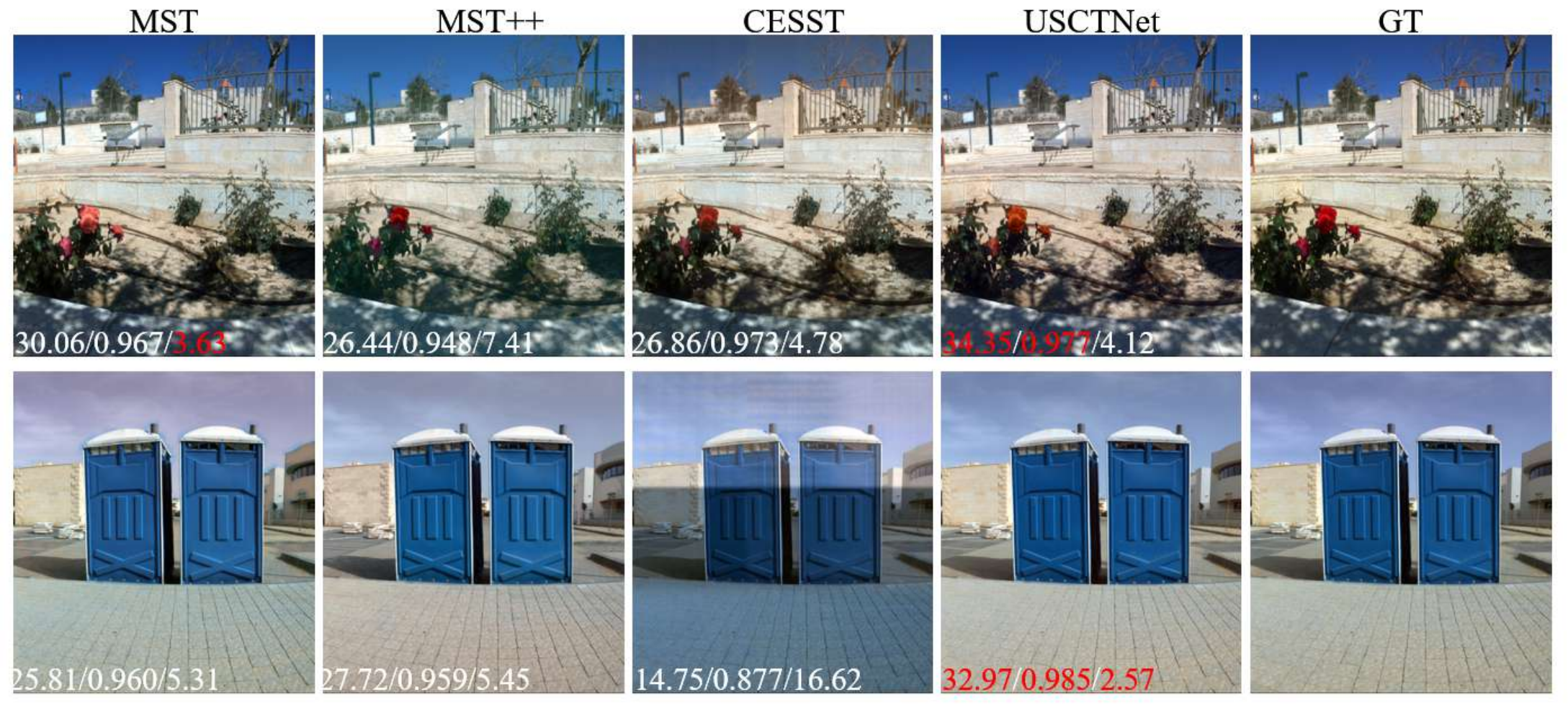}
    \caption{Reproduced RGBs from reconstructed HSIs with PSNR, SSIM and $\Delta E_{00}$ metrics.}
    \label{fig.4}
\end{figure}

\textbf{Qualitative Results.} Fig.~\ref{fig.2} visualizes mean-squared-error maps across spectral bands. Fig.~\ref{fig.3} presents reconstructions at 550, 600, 650, and 700\,nm, where USCTNet recovers finer spatial details while preserving spectral consistency. Fig.~\ref{fig.4} compares RGB reproduction, we also report the color difference consistency of RGB reproduction evaluated by $\Delta E_{00}$. Our outputs closely match the ground truth and attain high PSNR/SSIM with relatively low $\Delta E_{00}$, indicating strong physical consistency.

\textbf{Ablation Studies.} We assess the contributions of the gradient-descent/proximal design (GD+prox), CAB, DRTB, and LRSP on ARAD-1K Real(Table~\ref{tab:ablation}) and we use MST++ as the baseline. Integrating the original SVT fails to improve accuracy due to the cost of full SVDs, increased gradient noise, and numerical instability of singular-value spectra in high dimensions. LRSP replaces it with learnable soft thresholding in a low-dimensional subspace, explicitly controlling the effective rank and smoothing gradient flow, which improves metrics under identical settings.

We also ablate the number of unfolding iterations $k$. The model fails to converge for $k<3$ and overfits for $k\!\ge\!5$. Although $k\!=\!4$ markedly increases parameters with only marginal gains over $k\!=\!3$, we set $k\!=\!3$ to balance convergence, generalization, and complexity.
\begin{table}
\centering
\renewcommand{\arraystretch}{1.2}
\setlength{\tabcolsep}{6pt}
\resizebox{0.9\linewidth}{!}{%
\begin{tabular}{l|c|c|c|c|c|c}
\hline
Method    & GD+prox & CAB & DRTB & LRSP & PSNR   & SAM   \\
\hline
Baseline  &         &     &      &      & 34.36  & 5.01   \\
Variant 1 & \checkmark &  &      &      & 33.21  & 5.68   \\
Variant 2 & \checkmark & \checkmark & &      &  34.07  & 5.24   \\
Variant 3 & \checkmark & \checkmark & \checkmark & & 35.15  & 4.71   \\
USCTNet   & \checkmark & \checkmark & \checkmark & \checkmark & \textbf{36.56} & \textbf{4.69} \\
\hline
\end{tabular}%
}
\caption{Ablation study of core modules.}
\label{tab:ablation}
\end{table}
\section{Conclusion}
We presented USCTNet, a physics-grounded framework for RGB-to-HSI reconstruction that reconciles spectral accuracy with RGB fidelity. USCTNet integrates imaging physics and structural priors through two key components: (1) explicit estimation of camera spectral sensitivity and illumination, coupled with a gradient-descent data-fidelity update for color-consistent reproduction; (2) a low-rank subspace proximal operator that enforces a nuclear-norm prior via efficient subspace SVT, improving gradient stability and scalability. Realized as an unfolding solver in a learnable transform domain, USCTNet achieves SOTA performance on standard benchmarks and substantially reduces RGB–HSI mismatches. By uniting data-driven modeling with physics, it provides a foundation for reliable spectral reconstruction in remote sensing and computational photography.




\begin{thebibliography}{99}

\bibitem{yang2024hyperspectral}
X.~Yang, J.~Chen, and Z.~Yang, ``Hyperspectral image reconstruction via combinatorial embedding of cross-channel spatio-spectral clues,'' in \emph{Proc. AAAI Conf. Artif. Intell.}, vol.~38, no.~7, pp. 6567--6575, 2024.

\bibitem{huo2024learning}
D.~Huo, J.~Wang, Y.~Qian, and Y.-H.~Yang, ``Learning to recover spectral reflectance from RGB images,'' \emph{IEEE Trans. Image Process.}, vol.~33, pp. 3174--3186, 2024.

\bibitem{lin2020physically}
Y.-T.~Lin and G.~D.~Finlayson, ``Physically plausible spectral reconstruction from RGB images,'' in \emph{Proc. CVPR Workshops}, pp. 532--533, 2020.

\bibitem{wang2017hyperspectral}
Y.~Wang, J.~Peng, Q.~Zhao, Y.~Leung, X.-L.~Zhao, and D.~Meng, ``Hyperspectral image restoration via total variation regularized low-rank tensor decomposition,'' \emph{IEEE J. Sel. Topics Appl. Earth Observ. Remote Sens.}, vol.~11, no.~4, pp. 1227--1243, 2017.

\bibitem{zhang2024t2lr}
Y.~Zhang, P.~Li, and Y.~Hu, ``T2LR-Net: An unrolling network learning transformed tensor low-rank prior for dynamic MR image reconstruction,'' \emph{Comput. Biol. Med.}, vol.~170, p.~108034, 2024.

\bibitem{zhang2024differentiable}
Y.~Zhang and Y.~Hu, ``Differentiable SVD based on Moore--Penrose pseudoinverse for inverse imaging problems,'' \emph{arXiv preprint arXiv:2411.14141}, 2024.

\bibitem{wang2024ufc}
X.~Wang and H.~Gan, ``UFC-Net: Unrolling fixed-point continuous network for deep compressive sensing,'' in \emph{Proc. CVPR}, pp. 25149--25159, 2024.

\bibitem{cai2022mst++}
Y.~Cai, J.~Lin, Z.~Lin, H.~Wang, Y.~Zhang, H.~Pfister, R.~Timofte, and L.~Van~Gool, ``MST++: Multi-stage spectral-wise transformer for efficient spectral reconstruction,'' in \emph{Proc. CVPR}, pp. 745--755, 2022.

\bibitem{ntiredataset}
B.~Arad, R.~Timofte, R.~Yahel, N.~Morag, A.~Bernat, Y.~Cai, J.~Lin, Z.~Lin, H.~Wang, Y.~Zhang, \emph{et~al.}, ``NTIRE 2022 spectral recovery challenge and data set,'' in \emph{Proc. CVPR}, pp. 863--881, 2022.

\bibitem{icvldataset}
B.~Arad and O.~Ben-Shahar, ``Sparse recovery of hyperspectral signal from natural RGB images,'' in \emph{Proc. ECCV}, pp. 19--34, 2016.

\bibitem{hsi2rgb}
M.~Magnusson, J.~Sigurdsson, S.~E.~Armansson, M.~O.~Ulfarsson, H.~Deborah, and J.~R.~Sveinsson, ``Creating RGB images from hyperspectral images using a color matching function,'' in \emph{Proc. IGARSS}, pp. 2045--2048, 2020.

\bibitem{cai2022mask}
Y.~Cai, J.~Lin, X.~Hu, H.~Wang, X.~Yuan, Y.~Zhang, R.~Timofte, and L.~Van~Gool, ``Mask-guided spectral-wise transformer for efficient hyperspectral image reconstruction,'' in \emph{Proc. CVPR}, pp. 17502--17511, 2022.

\bibitem{li2020adaptive}
J.~Li, C.~Wu, R.~Song, Y.~Li, and F.~Liu, ``Adaptive weighted attention network with camera spectral sensitivity prior for spectral reconstruction from RGB images,'' in \emph{Proc. CVPR Workshops}, pp. 462--463, 2020.

\bibitem{chen2021hinet}
L.~Chen, X.~Lu, J.~Zhang, X.~Chu, and C.~Chen, ``HINet: Half instance normalization network for image restoration,'' in \emph{Proc. CVPR}, pp. 182--192, 2021.

\bibitem{zamir2021multi}
S.~W.~Zamir, A.~Arora, S.~Khan, M.~Hayat, F.~S.~Khan, M.-H.~Yang, and L.~Shao, ``Multi-stage progressive image restoration,'' in \emph{Proc. CVPR}, pp. 14821--14831, 2021.

\bibitem{zhang2018ista}
J.~Zhang and B.~Ghanem, ``ISTA-Net: Interpretable optimization-inspired deep network for image compressive sensing,'' in \emph{Proc. CVPR}, pp. 1828--1837, 2018.

\bibitem{halko2011RandQB}
N.~Halko, P.-G.~Martinsson, and J.~A.~Tropp, ``Finding structure with randomness: Probabilistic algorithms for constructing approximate matrix decompositions,'' \emph{SIAM Rev.}, vol.~53, no.~2, pp. 217--288, 2011.

\bibitem{xu2017low}
Y.~Xu, Z.~Wu, J.~Chanussot, M.~Dalla~Mura, A.~L.~Bertozzi, and Z.~Wei, ``Low-rank decomposition and total variation regularization of hyperspectral video sequences,'' \emph{IEEE Trans. Geosci. Remote Sens.}, vol.~56, no.~3, pp. 1680--1694, 2017.

\bibitem{zamir2022restormer}
S.~W.~Zamir, A.~Arora, S.~Khan, M.~Hayat, F.~S.~Khan, M.-H.~Yang, and L.~Shao, ``Restormer: Efficient transformer for high-resolution image restoration,'' in \emph{Proc. CVPR}, pp. 5728--5739, 2022.

\bibitem{zhang2025jotlasnet}
Y.~Zhang, H.~Gui, N.~Yang, and Y.~Hu, ``JotlasNet: Joint tensor low-rank and attention-based sparse unrolling network for accelerating dynamic MRI,'' \emph{Magn. Reson. Imaging}, vol.~118, p.~110337, 2025.

\bibitem{xiao2017svd}
Y.~Xiao, Z.~Li, T.~Yang, and L.~Zhang, ``SVD-free convex--concave approaches for nuclear norm regularization,'' in \emph{Proc. IJCAI}, pp. 3126--3132, 2017.

\end{thebibliography}
\end{document}